\documentclass{bmvc2k}

\usepackage{graphicx}
\usepackage{cleveref}
\usepackage{multirow}
\usepackage{algorithm}
\usepackage{algpseudocode}
\usepackage{booktabs}
\usepackage{wrapfig}
\usepackage{enumitem}

\title{Trimming the Fat: Efficient Compression of 3D Gaussian Splats through Pruning}

\addauthor{Muhammad Salman Ali}{salmanali@khu.ac.kr}{1,2}
\addauthor{Maryam Qamar}{maryamqamar@khu.ac.kr}{2}
\addauthor{Sung-Ho Bae}{shbae@khu.ac.kr}{2}
\addauthor{Enzo Tartaglione}{enzo.tartaglione@telecom-paris.fr}{1}

\addinstitution{
 LTCI, Télécom Paris,\\ Institut Polytechnique de Paris, France
}
\addinstitution{
 Kyung Hee University,\\ Republic of Korea
}

\runninghead{Ali et al.}{Efficient Compression of 3DGS through Pruning}

\def\etal{\emph{et al}\bmvaOneDot}

\begin{document}

\maketitle
\begin{abstract}
In recent times, the utilization of 3D models has gained traction, owing to the capacity for end-to-end training initially offered by Neural Radiance Fields and more recently by 3D Gaussian Splatting (3DGS) models. The latter holds a significant advantage by inherently easing rapid convergence during training and offering extensive editability. However, despite rapid advancements, the literature still lives in its infancy regarding the scalability of these models. In this study, we take some initial steps in addressing this gap, showing an approach that enables both the memory and computational scalability of such models. Specifically, we propose ``Trimming the fat'', a post-hoc gradient-informed iterative pruning technique to eliminate redundant information encoded in the model. 
Our experimental findings on widely acknowledged benchmarks attest to the effectiveness of our approach, revealing that up to 75\% of the Gaussians can be removed while maintaining or even improving upon baseline performance. 
Our approach achieves around 50$\times$ compression while preserving performance similar to the baseline model, and is able to speed-up computation up to 600~FPS. The code can be found \href{https://github.com/salmanali96/Trimming-the-Fat}{here}.
\end{abstract}

\section{Introduction}
\label{sec:intro}
\begin{figure}[t]
    \centering
    \includegraphics[width=0.7\linewidth]{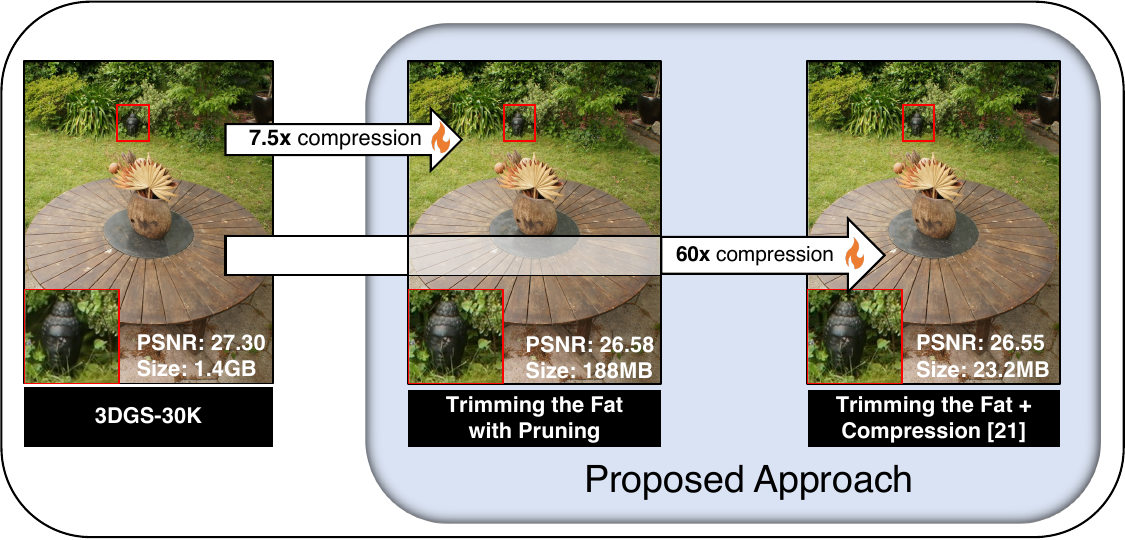}
    \caption{
    Vanilla 3DGS-30k vs our novel pruning approach applied with an end-to-end compression technique~\cite{niedermayr2023compressed}.
    }
    \label{fig:teaser}
\end{figure}
In the last few years, significant advancements have been made in radiance field methodologies for reconstructing 3D scenes using images captured from various viewpoints. The emergence of Neural Radiance Fields (NeRF) techniques has notably influenced the realm of 3D scene modeling and reconstruction~\cite{mildenhall2021nerf,huang2022stylizednerf}. The efficient generation of photo-realistic novel views from a given set of training images has become a focal point in computer vision research, with diverse applications~\cite{deng2023compressing}. NeRF's capability to distill the essence of a 3D object from its 2D representations, while maintaining compactness, underscores its impact and popularity in the literature~\cite{yu2021pixelnerf,kaya2022neural}.

Despite its success, the traditional NeRF~\cite{mildenhall2021nerf} suffers from slow training and rendering speeds. To address this challenge, various approaches have been proposed, although they often entail compromises in rendered image quality~\cite{deng2023compressing}. Recent studies have turned to explicit scene representations, such as voxel-based~\cite{sun2022direct} or point-based~\cite{xu2022point} structures to enhance rendering efficiency. For instance, leveraging 3D voxel grids on GPUs alongside multi-resolution hash encoding of inputs led to consistent reductions in required operations and enabled real-time performance~\cite{muller2022instant}. Similarly, the most efficient radiance field solutions to date rely on continuous representations achieved through interpolating values stored in voxel~\cite{fridovich2022plenoxels}, hash grids~\cite{muller2022instant}, or points~\cite{xu2022point}. While the continuous nature of these methods aids optimization, the stochastic sampling necessary for rendering can incur computational overhead and introduce noise~\cite{kerbl20233d}.


A recent advancement in the field is the introduction of differentiable 3D Gaussian splatting (3DGS), which enables the generation of a sparse adaptive scene representation~\cite{kerbl20233d}. This representation can be rendered rapidly on the GPU, offering big speed improvements. 3DGS combines the best features of existing methods: leveraging a 3D Gaussian representation for scene optimization provides state-of-the-art visual quality and competitive training times while the tile-based splatting solution ensures real-time rendering at high quality for 1080p resolution across various datasets. Unlike NeRF methods, 3DGS simplifies training and rendering by projecting 3D Gaussians to the 2D image space and combines them with opacity using \emph{rasterization}, enabling real-time rendering on a single GPU. Furthermore, the explicit storage of scene structure in the parameter space allows for direct editing of the 3D scene. However, some challenges emerge when employing differentiable 3DGS, particularly in optimizing scenes with millions of Gaussians, which may require substantial storage and memory. While specialized pipelines demonstrate real-time performance on high-end GPUs, seamless integration into VR/AR environments or games remains a challenge, particularly when working alongside hardware rasterization of polygon models.

In this paper, we aim to compress Gaussian splatting representations while preserving their rendering speed and quality, facilitating their application across diverse domains such as IoT devices with limited storage or memory. Our primary insight is that the learned 3DGS models exhibit over-fitting to the underlying scene, allowing for the removal or pruning of many Gaussians without sacrificing performance, particularly due to markedly lower opacity values. We start the training process with a pre-trained optimized Gaussian scene, iteratively pruning it based on opacity levels and gradient values, followed by fine-tuning to achieve superior performance-compression trade-off compared to the baseline optimized scene as showcased from Fig.~\ref{fig:teaser}. Our main contributions are the following.


\begin{itemize}[noitemsep, nolistsep]
    \item We build on top of the optimized 3DGS as a 3D prior for pruning, enabling the removal of redundant Gaussians while fine-tuning the remaining ones to accurately capture the scene features (Sec.~\ref{sec: pruning}). 
    
    \item We observe that vanilla pruning is sub-optimal when compared to a gradient-informed approach and that pruning without such a prior fails. Besides, we showcase the compatibility with other compression pipelines, like~\cite{niedermayr2023compressed} (Sec.~\ref{sec:abl}).

    \item With our proposed method, we achieve state-of-the-art performance even after pruning 50\% of the Gaussian splats, significantly enhancing the scalability of 3DGS (Sec.~\ref{sec:res}). Our compression pipeline achieves an enhanced balance between scene fidelity and compression, surpassing the baseline (Fig.~\ref{fig: Pruning-Comparison-1}). 
\end{itemize}


\section{Related Work}
\label{sec:Related Work}
In this section we first provide an overview of the most recent methods for novel view synthesis (Sec.~\ref{sec:nws}), then we discuss approaches for their compression (Sec.~\ref{sec:3dgs-compress}).

\subsection{Novel View Synthesis}
\label{sec:nws}
Recent advancements in novel view synthesis have seen significant progress, with early techniques using CNNs to estimate blending weights or texture-space solutions~\cite{flynn2016deepstereo, hedman2018deep,riegler2020free,thies2019deferred}, albeit facing challenges with MVS-based geometry and temporal flickering. Volumetric representations, starting with Soft3D~\cite{penner2017soft} and employing deep learning with volumetric ray-marching~\cite{henzler2019escaping,sitzmann2019deepvoxels}, provided further advancements. Neural Radiance Fields (NeRFs)~\cite{mildenhall2021nerf} aimed to enhance synthesized views' quality but faced slow processing due to a large Multi-Layer Perceptron (MLP) backbone and dense sampling. Subsequent methods like Mip-NeRF360~\cite{barron2022mip} focused on balancing quality and speed, while recent advances prioritize faster training and rendering through spatial data structures, encodings, and MLP adjustments~\cite{chen2023mobilenerf,fridovich2022plenoxels,garbin2021fastnerf}. Notable methods such as InstantNGP~\cite{muller2022instant} leverage hash grids and occupancy grids for accelerated computation, while Plenoxels~\cite{fridovich2022plenoxels} rely on Spherical Harmonics for directional effects without neural networks. Despite these strides, challenges persist in NeRF methods regarding efficient coding for empty space, image quality, and rendering speed.

In contrast, 3DGS achieves superior quality and faster rendering without implicit learning. However, its increased storage compared to NeRF methods poses limitations. Our approach aims to maintain the quality and speed of 3DGS while reducing model storage by applying pruning to Gaussian parameters.
\begin{figure*}[t]
    \centering
    \includegraphics[width=0.8\linewidth]{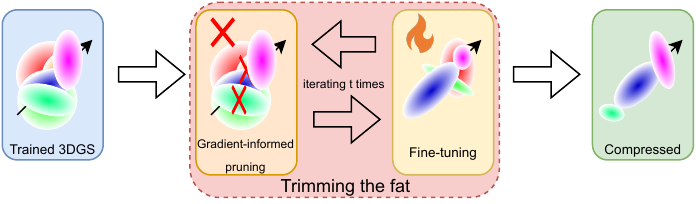}
    \caption{Overview of our pruning pipeline. From a pre-trained 3DGS-30k scene, we first iteratively prune it for a fixed number of iterations with subsequent fine-tuning. Then, we conduct further fine-tuning for 20,000 iterations to obtain our final optimized scene.}
    \label{fig:overview}
\end{figure*}
\subsection{3DGS Compression}
\label{sec:3dgs-compress}
When compared to NeRFs, 3DGS models lack structure, which presents challenges for compression~\cite{chen2024hac,fei20243d}. Consequently, many studies in 3DGS compression introduce structural parameters by replacing vanilla 3DGS parameters to enhance compression~\cite{lu2023scaffold, chen2024hac}. Scaffold-GS~\cite{lu2023scaffold}, for instance, utilizes anchor points to distribute local 3D Gaussians and predicts their attributes dynamically based on the viewing direction and distance within the view frustum. On the other hand, the Hash-grid Assisted Context (HAC)~\cite{chen2024hac} framework jointly learns a structured compact hash grid and uses it for context modeling of anchor attributes. 

Niedermayr~\etal~\cite{niedermayr2023compressed} proposed a compression framework that maintains vanilla 3DGS parameters while compressing directional colors and Gaussian parameters. This framework incorporates sensitivity-aware vector clustering and quantization-aware training and achieves compression rates of up to 30$\times$ with a marginal decline in performance compared to the baseline 3DGS. Another method, Compact3D~\cite{lee2023compact}, introduces a learnable mask strategy to prune the Gaussians and a compact representation of view-dependent colors by employing a grid-based neural field rather than relying on spherical harmonics. It also learns codebooks to compactly represent the geometric attributes of Gaussian by vector quantization.

Our proposed pruning method can effectively improve/replace the masking strategies for unimportant Gaussians in the existing works (see Sec.~\ref{sec:res}). However, for the scope of this paper and considering the broad applicability of vanilla 3DGS, we focus solely on the vanilla 3DGS variant.


\section{Methodology}
In this section, we first provide an overview of the 3DGS technique for learning and rendering 3D scenes, as introduced by Kerbl~\etal~\cite{kerbl20233d} (Sec.~\ref{sec: 3DGS}). Then, we delve into an explanation of our pruning approach (Sec.~\ref{sec: pruning}). Our methodology, depicted in Fig.~\ref{fig:overview}, introduces an effective approach to compress these models using gradient-informed pruning.

\subsection{Differentiable Gaussian Splatting}
\label{sec: 3DGS}
3DGS~\cite{kerbl20233d} represents a scene through a set of 3D Gaussians. By leveraging differentiable Gaussian splatting, which extends EWA volume splatting~\cite{zwicker2001ewa}, it facilitates the efficient projection of 3D Gaussian kernels onto the 2D image plane. Additionally, differentiable rendering optimizes the quantity and attributes of the Gaussian kernels employed to characterize the scene. Each 3D Gaussian is characterized by its position and covariance matrices within the 3D space, modeled as
\begin{equation}
G(x)=\exp\left[-\frac{1}{2}(x-\mu)^T \Sigma^{-1}(x-\mu)\right],
\end{equation}

where \(x\) denotes the position vector, \(\mu\) represents the position, and \(\Sigma\) is the 3D covariance matrix of the Gaussian distribution. Given the requirement for the covariance matrix to be positive definite, it can be parameterized using a rotation matrix \(R\) and a scaling matrix \(S\). To facilitate independent optimization of \(R\) and \(S\), Kerbl~\etal~\cite{kerbl20233d} introduce a representation of rotation via a quaternion \(q\) and scaling through a vector \(s\), both of which can be converted into their corresponding matrices. Additionally, each Gaussian distribution possesses its opacity (\(\alpha \in [0, 1]\)) and a set of spherical harmonics (SH) coefficients essential for reconstructing a view-dependent color. The 2D projection of a 3D Gaussian remains a Gaussian with covariance
\begin{equation}
    \Sigma' = JW\Sigma W^TJ^T,
\end{equation}
\noindent
where \(W\) is the view transformation matrix, and \(J\) is the Jacobian of the affine approximation of the projective transformation. This setup facilitates the evaluation of the 2D color and opacity footprint of each projected Gaussian. The color \(C\) of a pixel is subsequently determined by blending all the \(N\) 2D Gaussians contributing to that pixel:
\begin{equation}
    C = \sum_{i \in N} c_i \alpha_i + \prod_{j=1}^{i-1} (1 - \alpha_j),
\end{equation}

\noindent
where \(c_i\) and \(\alpha_i\) represent the view-dependent color and opacity of a Gaussian, respectively, which are adjusted based on the exponential decay from the center point of the projected Gaussian. The parameters such as position \(x\), rotation \(q\), scaling \(s\), opacity \(\alpha\), and spherical harmonics (SH) coefficients of each 3D Gaussian are optimized to ensure alignment between the rendered 2D Gaussians and the training images.

During the training phase, the 3D Gaussian splats are rendered efficiently in a differentiable manner to produce a 2D image. This rendering process involves $\alpha$-blending of anisotropic splats, sorting them, and utilizing a tile-based rasterizer. At each training iteration, the 3DGS framework renders the training viewpoints and then minimizes the loss between the ground truth and rendered images in the pixel space, where the loss is given by 
\begin{equation}
\mathcal{L}=(1-\lambda) \mathcal{L}_1+\lambda \mathcal{L}_{\text {D-SSIM }},
\end{equation}

\noindent
where $\mathcal{L}_1$ is the $\ell_1$ norm of the rendered output and $\mathcal{L}_{\text {D-SSIM }}$ is its structural dissimilarity. The optimization in 3DGS begins with a point cloud generated through a conventional SfM method~\cite{ullman1979interpretation}, and then proceeds iteratively, pruning Gaussians with small opacity parameters and introducing new ones when significant gradients are detected. As demonstrated in the 3DGS paper, this approach enables rapid training and facilitates real-time rendering, all while achieving comparable or superior 3D model quality compared to state-of-the-art NeRF methods.

\subsection{Gradient Aware Pruning}
\label{sec: pruning}
3DGS typically necessitates several million Gaussians to adequately model a standard scene, each Gaussian entailing 59 parameters. This results in a storage size significantly larger than that of most NeRF methodologies, such as Mip-NeRF360~\cite{barron2022mip}, K-planes~\cite{fridovich2023k}, and InstantNGP~\cite{muller2022instant}. Such requirements render 3DGS inefficient for certain applications, particularly those involving edge devices. Our primary focus is on parameter reduction. In the original training process of 3DGS, Kerbl~\etal~\cite{kerbl20233d} pruned and densified Gaussians up to a specified number of iterations, with pruning based on a predetermined opacity threshold. However, if their opacity is low and so is its gradient, we can say they can be removed with little to no impact on the quality of the rendered scene. As such, we have 


    

    
\begin{equation}
    \label{eq: pruning}
    \boldsymbol{\Sigma}' = \begin{cases}
     \Sigma_i & \text{if } \left[\left|\Sigma_i^\alpha\right|\geq \mathcal{Q}_{|\boldsymbol{\Sigma^\alpha}|}(\gamma_{\text{iter}})\right] \lor    \left[ \left|\nabla\Sigma_i\right| \geq \mathcal{Q}_{|\boldsymbol{\nabla\Sigma}|}(\gamma_{\text{iter}}) \right] \\
    0  & \text{otherwise,}
    \end{cases}
\end{equation}

\noindent where $\Sigma_i^\alpha$ denotes the $\alpha$ value of the $i$-th Gaussian, and $\nabla\Sigma_i$ denotes the gradient for the $i$-th Gaussian, $\mathcal{Q}_{|\boldsymbol{\Sigma^\alpha}|}(.)$ represents the quantile function for the opacity, $\mathcal{Q}_{|\boldsymbol{\nabla\Sigma}|}(.)$ is the quantile function for the gradients of the Gaussians, and $\gamma \in [0, 1]$ denotes the fraction of Gaussians to be removed. 
This pruning process, along with periodic fine-tuning, not only improves performance but evidently results in substantial compression gains.



The iterative pruning and fine-tuning approaches enable the removal of redundant Gaussians while refining the remaining ones to better capture scene details compared to the baseline. Prior research in scene rendering has demonstrated that a gradual iterative pruning strategy can yield significantly sparser models while preserving high fidelity~\cite{deng2023compressing}. However, the impact of such an approach on 3DGS models yet remains unclear. We speculate that by gradually pruning the model over a specified number of iterations $t$ and aiming for a target sparsity $\gamma_{\text{target}}$, we can achieve improved results through a sparsification process applied to the 3DGS model: at every iteration, we apply the sparsification
\begin{equation}
    \gamma_{\text{iter}} = 1 - (1-\gamma_{\text{target}})^{\frac{1}{t}}.
\end{equation}
\noindent Our approach is based on two key factors. First, during the fine-tuning stage following pruning, the covariance $\Sigma$ is adjusted to minimize rendering loss, leading to higher values for solid surfaces and lower values for semi-transparent artifacts, which can be subsequently removed in the next iteration (which will be empirically observed in Fig.~\ref{fig: iterative vs one-shot}c). Second, the gradual iterative process helps prevent the optimization algorithm from converging to sub-optimal local minima (according to empirical evidence discussed in Sec.~\ref{sec:abl}). Hence, it is crucial to initiate the process with an overparametrized yet well-performing model. Similar observations can be drawn from traditional deep learning literature~\cite{blalock2020state, woodworth2020kernel}. In the next section, we will present a quantitative analysis of typical benchmarks employed for 3DGS.

\begin{figure*}[t]
    \centering
    \includegraphics[width=1.0\linewidth]{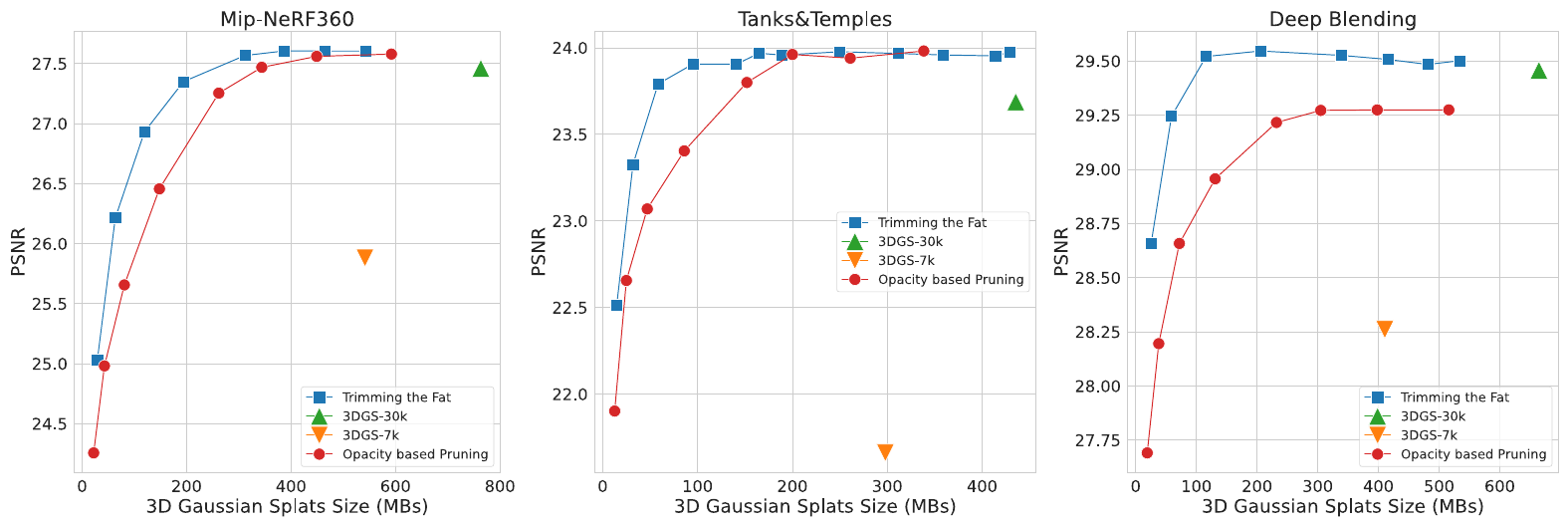}
    \caption{The graph depicts the trade-off between performance and size when utilizing Trimming the Fat (gradient-aware iterative pruning) compared to the 3DGS-30k and 3DGS-7k baselines as well as opacity-based pruning on the Mip-NeRF360, Tanks\&Temple, and Deep Blending datasets.}
    \label{fig: comparison}
\end{figure*}

\begin{figure*}[t]
\includegraphics[width=0.9\linewidth]{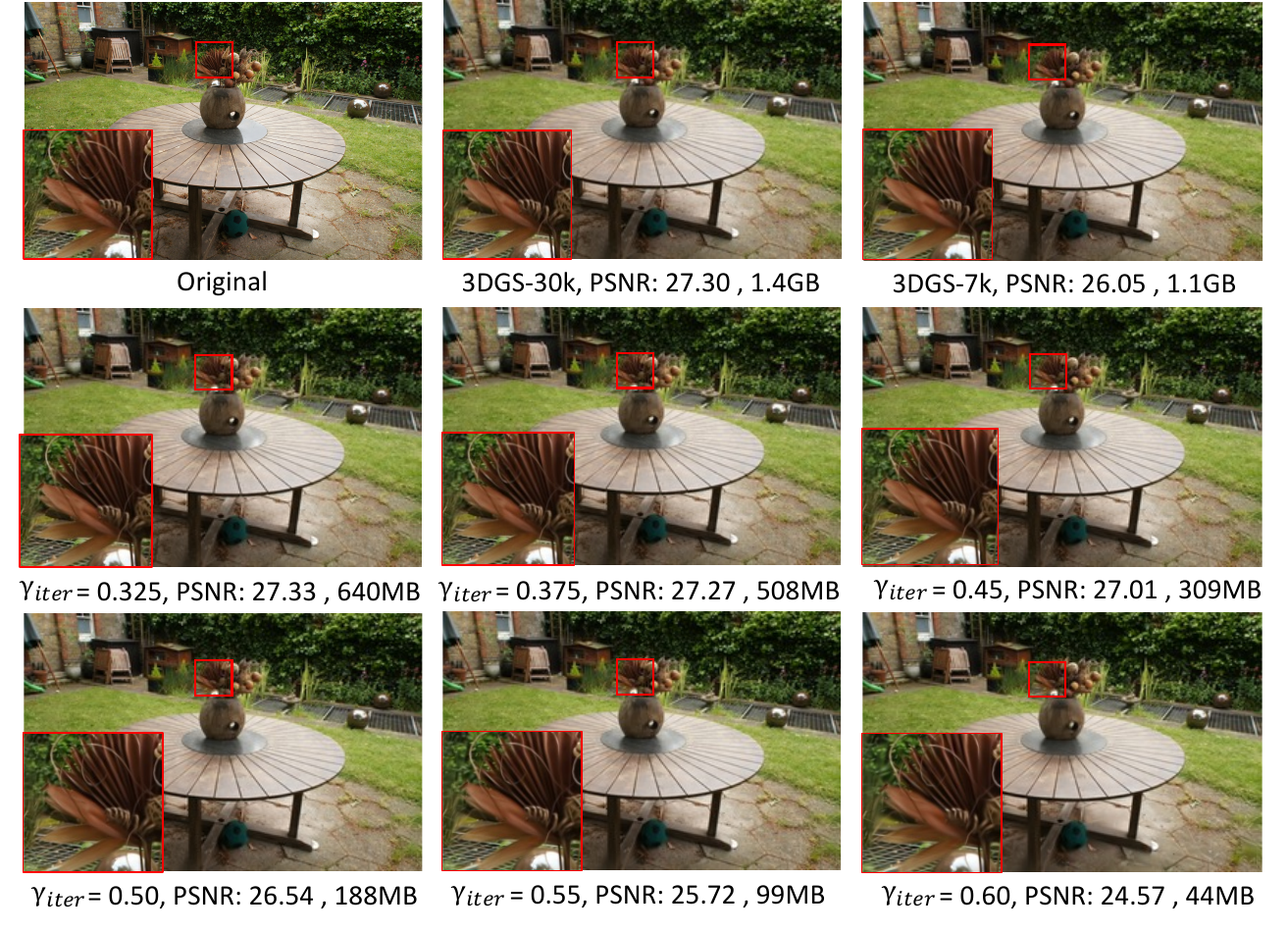}
    \centering
    \caption{Qualitative comparison of the \texttt{garden} scene at various pruning levels ($\gamma_{\text{iter}}$) using Trimming the Fat (gradient-aware iterative pruning). Our proposed method demonstrates substantially higher compression rates compared to both baselines while maintaining similar visual quality.
    }
    \label{fig: Pruning-Comparison-1}
\end{figure*}

\section{Experiments and Results}
In this section, we present our empirical findings on commonly recognized benchmarks within the 3DGS community. We begin by detailing the implementation of our approach, followed by an outline of the benchmarked datasets and the evaluation metrics employed (Sec.~\ref{sec:idf}). Subsequently, we discuss both qualitative and quantitative results (Sec.~\ref{sec:res}), providing as well an ablation study (Sec.~\ref{sec:abl}).

\subsection{Implementation Details}
\label{sec:idf}
In all our experiments, we use the publicly available official code repository of 3DGS~\cite{kerbl20233d}, adhering to the recommended hyperparameter settings used for training to maintain consistency with the original 3DGS model. We initiate the pruning process with the optimized Gaussians trained for 30,000 iterations, employing pruning with 
$\gamma_{\text{iter}}\in[0.225,0.6]$, as described in \eqref{eq: pruning}.  Iterative pruning with the same $\gamma_{\text{iter}}$ value is applied after every 500 iterations until reaching 35,000 iterations (commencing from 30,000 iterations), followed by further fine-tuning for 10,000 iterations. For all our experiments, pruning and fine-tuning consistently yield significantly improved compression-performance trade-offs. Additionally, $\lambda= 0.2$ is employed consistently across all our experiments. 

\noindent\textbf{Datasets.} We assess the efficacy of our pruning approach across diverse scenes, encompassing environments from the Mip-Nerf360~\cite{barron2022mip} indoor and outdoor datasets, alongside two scenes sourced from the Tanks\&Temples~\cite{knapitsch2017tanks} and Deep Blending~\cite{hedman2018deep} datasets, akin to the scenes examined in the original 3DGS work~\cite{kerbl20233d}.

\noindent\textbf{Evaluation.} To ensure a fair comparison, we adhere to the same train-test split utilized in Mip-Nerf360~\cite{barron2022mip} and 3DGS~\cite{kerbl20233d}. 
Our evaluation encompasses standard metrics like SSIM, PSNR, and LPIPS, alongside the average memory consumption across all datasets.

\subsection{Results}
\label{sec:res}
\subsubsection{Quantitative Comparison} 
\textbf{Trimming the Fat.} We conduct a comparative analysis between our method, the 3DGS-30k, and 3DGS-7k baselines, along with an opacity-based pruning approach that removes gradient information from \eqref{eq: pruning}. As illustrated in Fig.~\ref{fig: comparison}, we examine the trade-off in compression performance across benchmark datasets. Across all the datasets, Gaussian splats can be pruned by up to 4$\times$, showcasing improved or similar performance compared to the baseline. Notably, even at significantly high pruning levels, where the average scene size is less than 25MB, our proposed pruning technique maintains comparable or even superior performance to that of the 3DGS-7k variant, achieving compression rates of up to 24$\times$ on average. This is realized without the need for any additional end-to-end compression pipeline integration, highlighting the standalone scalability of our proposed approach. 

Opacity-based pruning exhibits similar performance to gradient-aware pruning at small pruning thresholds. However, the performance difference becomes more pronounced at higher compression rates, as evident from the results in Fig.~\ref{fig: comparison}. Incorporating gradient information leads to additional performance improvements in pruning. This enhancement arises because certain scene features (sky, glass, etc.) may have low opacity but are still crucial for overall scene rendering. By considering gradient information, we ensure that only Gaussians containing unimportant features are pruned. Our proposal of incorporating information on the gradient shows its prominent effectiveness at higher pruning rates. For a more detailed comparison please refer to the supplementary material.

\begin{figure}
    \begin{tabular}{@{}c@{}c@{}c@{}}
        \bmvaHangBox{\includegraphics[width=0.3\columnwidth]{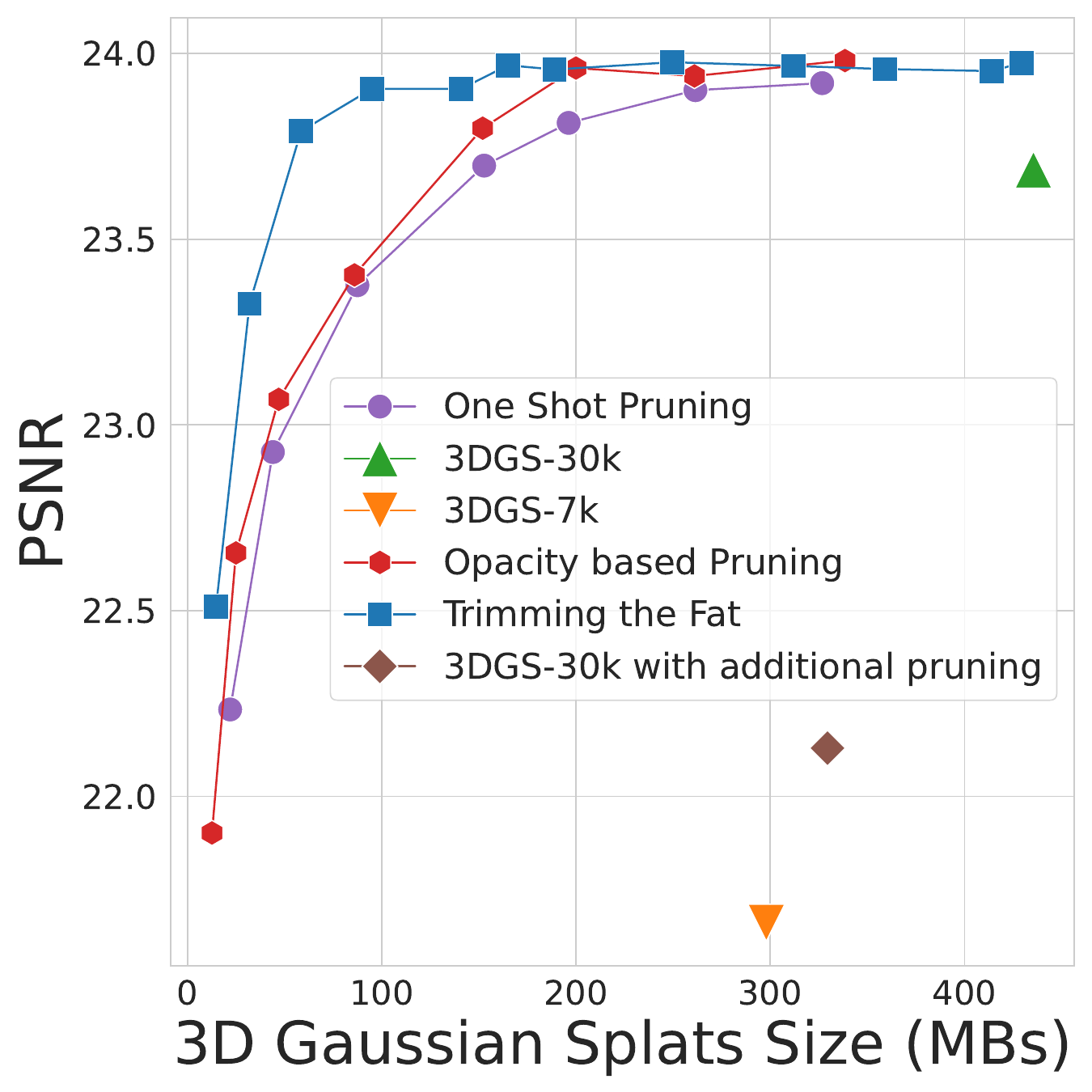}}&
        \bmvaHangBox{\includegraphics[width=0.3\columnwidth]{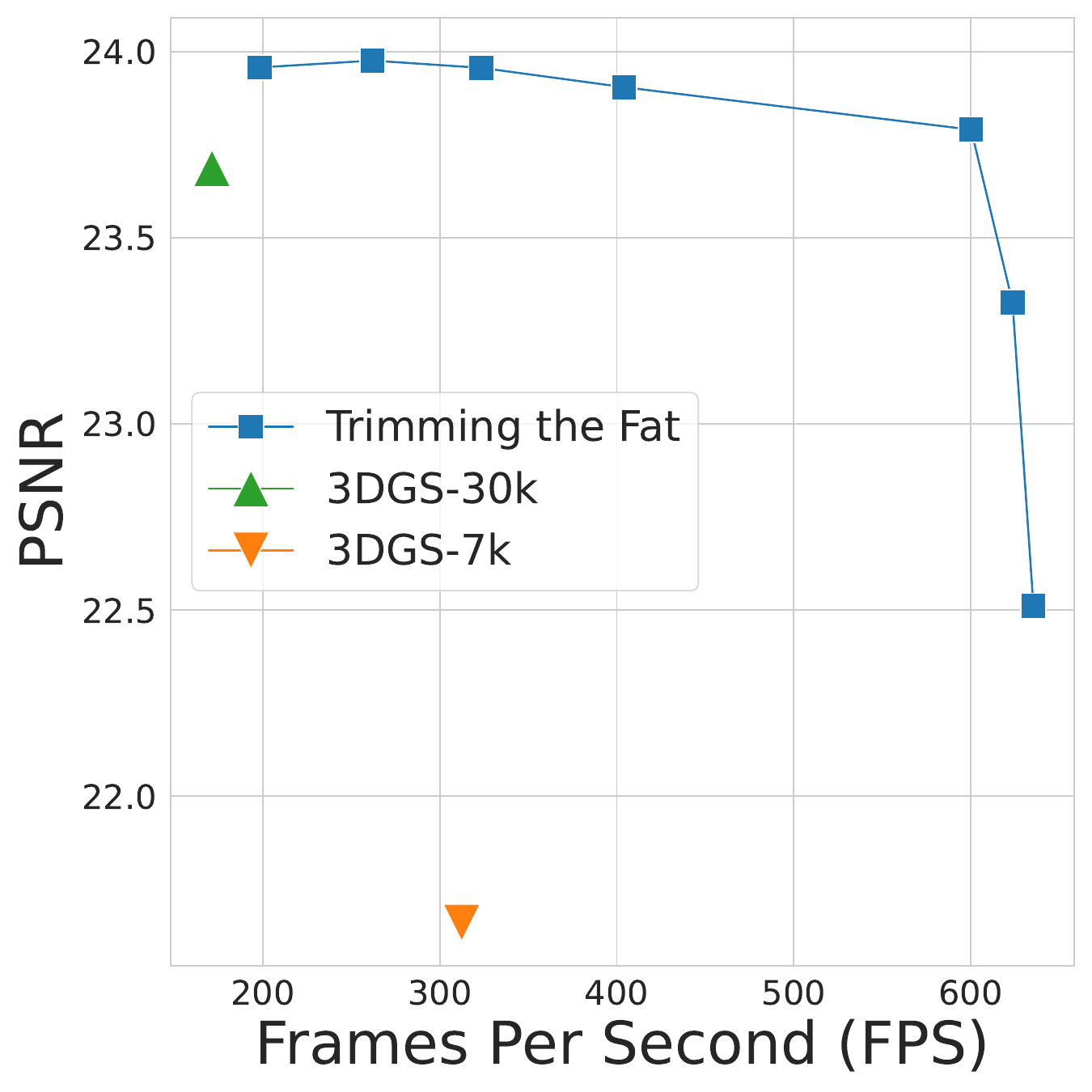}}&
        \bmvaHangBox{\includegraphics[width=0.4\columnwidth]{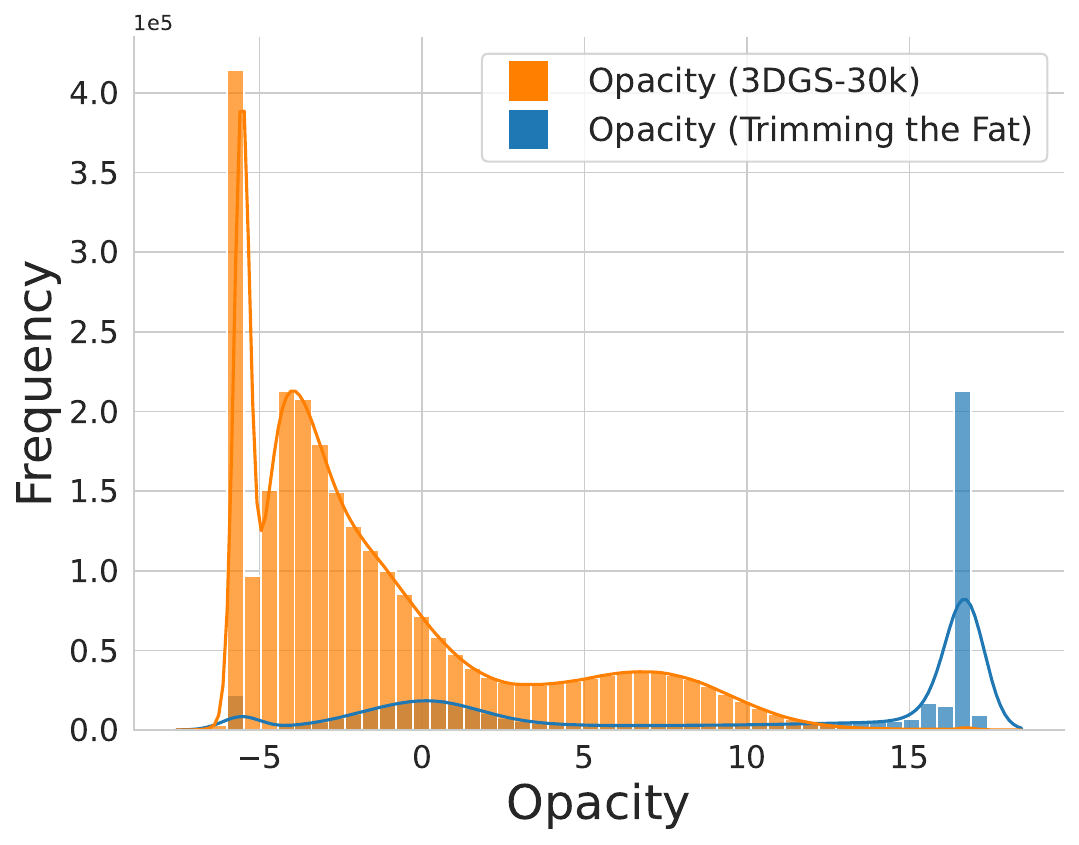}}\\
        (a)&(b)&(c)\\
    \end{tabular}
    \caption{Trade-off between performance and size through iterative pruning and one-shot pruning techniques (a) and in terms of FPS on the Tanks\&Temples dataset (b), and opacity distribution before and after pruning for the \texttt{truck} scene (c).}
    \label{fig: iterative vs one-shot}
\end{figure}

%
\noindent
\textbf{Trimming the Fat with end-to-end compression.}
Our proposed pruning methodology can act as a plug-and-play with various end-to-end compression techniques for 3DGS. When integrated with the method proposed by Niedermayr~\emph{et~al.}~\cite{niedermayr2023compressed}, we achieve state-of-the-art compression performance. Niedermayr's approach begins with a pre-trained Gaussian as the foundation of its compression process. We substitute this pre-trained Gaussian with our pruned Gaussian and apply the end-to-end compression procedure. This combination results in 50$\times$ compression compared to the baseline, while maintaining comparable performance. Moreover, we achieve 2$\times$ compression with improved performance compared to Niedermayr's original approach, as demonstrated in Table~\ref{tab:compression_comp}. 

\subsubsection{Qualitative Comparison} We present visualizations of the \texttt{garden} scene from the Mip-NeRF360 dataset, which typically demands substantial memory resources. In Fig.~\ref{fig: Pruning-Comparison-1} we showcase the visualizations of test set images at various pruning levels denoted by $\gamma_{\text{iter}}$. Our gradient-aware iterative pruning pipeline demonstrates significant compression rates while maintaining comparable visual quality. Specifically, our method compresses the Gaussian splats by approximately $4\times$ with visual quality akin to 3DGS-30K. Moreover, with $\gamma_{\text{iter}}=0.60$, our method achieves a compression ratio of approximately 25$\times$ while preserving visual quality comparable to 3DGS-7K. For additional visualizations, please refer to the supplementary material.


\subsection{Ablation Study}
\label{sec:abl}
\textbf{FPS Gain with Trimming the Fat.} Our novel pruning approach significantly enhances the FPS rate of 3DGS. On the Tanks\&Temples dataset, our method achieves an FPS of over 600+ all while maintaining SOTA performance as shown in~Fig.~\ref{fig: iterative vs one-shot}b. The renderings were performed using an RTX-3090, and the final FPS reported were averaged over three separate runs. These findings demonstrate the scalability of our proposed method.



\noindent\textbf{Why 3D Prior for Pruning is important?}
To assess the significance of a 3D prior in the pruning process, we modified the original training protocol introduced by Kerbl~\etal~\cite{kerbl20233d}. In their methodology, Gaussians are pruned and densified up to a specified iteration count (15k), employing an opacity threshold for pruning. Our modification involved halting the densification phase at the same iteration count (15k) but extending the pruning phase for an additional 10k iterations. Subsequently, the model underwent further fine-tuning for an additional 5k iterations to generate the final scene. The findings are outlined in Fig.~\ref{fig: iterative vs one-shot}a, indicating that even with a reduced pruning threshold, achieving convergence without a robust 3D prior remains challenging for a 3DGS model.

\noindent\textbf{Why Pruning is effective?}
Our proposed pruning technique achieves compression ratios of up to $4\times$ without compromising performance compared to the 3DGS baseline. The efficacy of our approach lies in its ability to effectively eliminate redundant Gaussians. As depicted in Fig.~\ref{fig: iterative vs one-shot}c, the opacity distribution before and after pruning for the \texttt{truck} changes significantly. For the baseline 3DGS, the majority of opacity values are very low, indicating minimal contribution to scene reconstruction. However, through post-hoc pruning, a significant proportion of opacity values become notably higher, indicating that a more solid geometry is learned by the model.

\noindent\textbf{One Shot Pruning vs Iterative Pruning.}
We also explored the impact of one-shot pruning in comparison to iterative pruning, in terms of model size, showcased in Fig.~\ref{fig: iterative vs one-shot}a. For one-shot pruning, we utilized the pre-trained 3DGS-30k model, performed the pruning process once, and then fine-tuned the model for 30k iterations. Both gradient-aware and opacity-based iterative pruning consistently outperformed the one-shot pruning method: our results demonstrate that gradual pruning enables the model to better adapt to the scene compared to one-shot pruning. 

\begin{table}[t]
    \caption{Performance and compression comparison of 3DGS baseline, end-to-end compression~\cite{niedermayr2023compressed}, and our proposed method with the reference end-to-end compression method. Comp. stands for compression rate. $^\dag$ Reported from~\cite{niedermayr2023compressed}. }
    \label{tab:compression_comp}
    \centering
    \resizebox{\textwidth}{!}{
    \begin{tabular}{l|ccccc|ccccc|ccccc}
    \toprule
      &  \multicolumn{5}{c}{3D Gaussian Splatting$^\dag$} & \multicolumn{5}{c}{End to End Compression~\cite{niedermayr2023compressed}$^\dag$} & \multicolumn{5}{c}{Our's Trimming the Fat with End to End Compression} \\
    \textbf{Dataset} & \textbf{SSIM$^\uparrow$} & \textbf{PSNR$^\uparrow$} & \textbf{LPIPS$^\downarrow$} & \textbf{Mem}$^\downarrow$ & \textbf{Comp.}$^\uparrow$ & \textbf{SSIM$^\uparrow$} & \textbf{PSNR$^\uparrow$} & \textbf{LPIPS$^\downarrow$} &  \textbf{Mem}$^\downarrow$ &  \textbf{Comp.}$^\uparrow$ & \textbf{SSIM$^\uparrow$} & \textbf{PSNR$^\uparrow$} & \textbf{LPIPS$^\downarrow$} & \textbf{Mem}$^\uparrow$ & \textbf{Comp.}$^\uparrow$\\
    \midrule
    
   Mip-NeRF360 & 0.812 & 27.287 & 0.220 & 795.263 & 1$\times$ &  0.801 & 26.981 & 0.238 & 28.803 & 26.230$\times$ & 0.798 & 27.130 & 0.248 & 20.057 & 39.650$\times$ \\
   Tanks\&Temples & 0.838  & 23.355 & 0.186 & 421.906 &   1$\times$ &  0.832 & 23.343 & 0.194 & 17.282 &  23.260$\times$  & 0.831 & 23.689 & 0.210 & 8.555 & 49.317$\times$ \\
   Deep Blending  & 0.898  & 29.432  & 0.246 & 703.772 & 1$\times$ & 0.898 & 29.381   & 0.253  & 25.299 & 27.816$\times$ &  0.897 & 29.425 & 0.267 & 12.494 & 56.329$\times$ \\

    \bottomrule
    \end{tabular}

    }
\end{table}





\section{Conclusion}
In this work, we presented a gradient-aware iterative pruning technique for 3D Gaussian splats named after ``Trimming the fat''. Our method effectively scales down Gaussian splats by a factor 4$\times$ without sacrificing generative quality. Particularly at higher pruning levels, our proposed method achieves compression ratios of approximately 25$\times$ and achieves up to 600~FPS with minimal impact on generative performance across established benchmark datasets. The resulting highly compressed point clouds can be seamlessly transmitted over networks and utilized on resource-constrained devices, offering potential applications in mobile VR/AR and gaming. Future research directions include investigating the integration of quantization-aware training methods to further improve the compressibility of 3DGS.

\section*{Acknowledgements}
This work was partially funded by Hi!PARIS Center on Data Analytics and Artificial Intelligence. This research was also supported in part by the MSIT (Ministry of Science and ICT), Korea, under the ITRC (Information Technology Research Center) support program (IITP-2023-RS-2023-00258649) supervised by the IITP(Institute for Information \& Communications Technology Planning \& Evaluation), and in part by the MSIT(Ministry of Science and ICT), Korea, under the ITRC (Information Technology Research Center) support program (IITP-2023-RS-2023-00259004) supervised by the IITP(Institute for Information \& Communications Technology Planning \& Evaluation).

\bibliography{main}

\title{Second Title}
\maketitle


\pagebreak

\begin{center}
 \section*{\large Supplementary Material for \\ Trimming the Fat: Efficient Compression of 3D Gaussian Splats through Pruning}
\end{center}

\section{Additional Results}

\subsection{Quantitative Comparison}
Table~\ref{tab:pruning_comp} provides an extensive comparison between our proposed approach and the 3DGS-30k and 3DGS-7k baseline methods across various $\gamma_{\text{iter}}$ values. Across all datasets, our method demonstrates the potential to prune Gaussian splats up to 4$\times$ while achieving performance improvements or maintaining comparable levels to the baseline. Even at highly compressed rates ($\gamma_{\text{iter}}$ = 0.6), our approach delivers reasonable performance similar to that of 3DGS-7k, while achieving average compression ratios of approximately 24$\times$.

\subsection{Qualitative Comparison}
\label{sec:intro}
\begin{figure*}
\includegraphics[width=\linewidth]{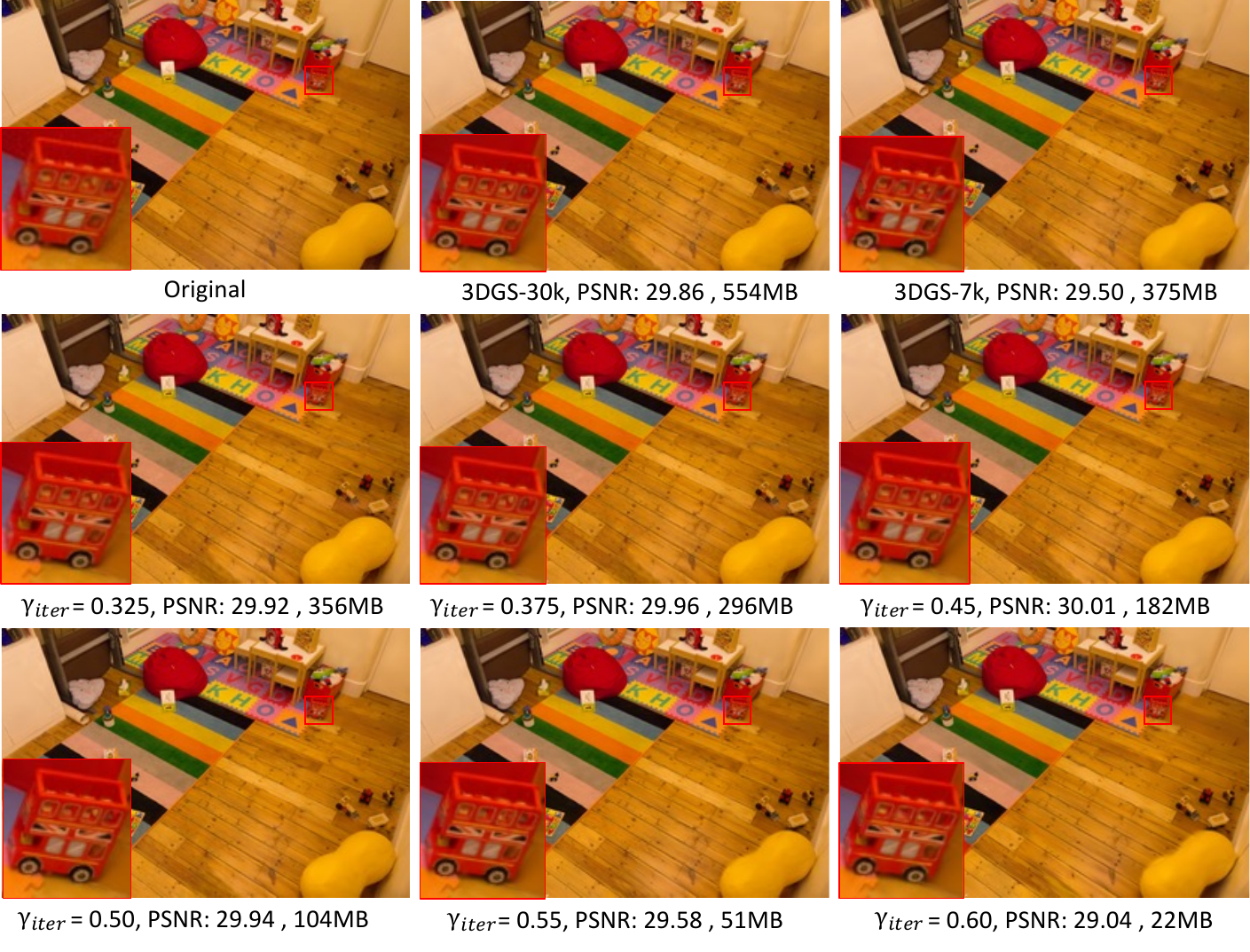}
    \centering
    \caption{Qualitative comparison of the playroom scene at various pruning levels, defined by $\gamma_{\text{iter}}$ using gradient-aware iterative pruning. Our proposed method demonstrates substantially higher compression rates compared to both baselines while maintaining similar visual quality.}
    \label{fig: Pruning-Comparison-2}
\end{figure*}

\begin{figure*}
\includegraphics[width=\linewidth]{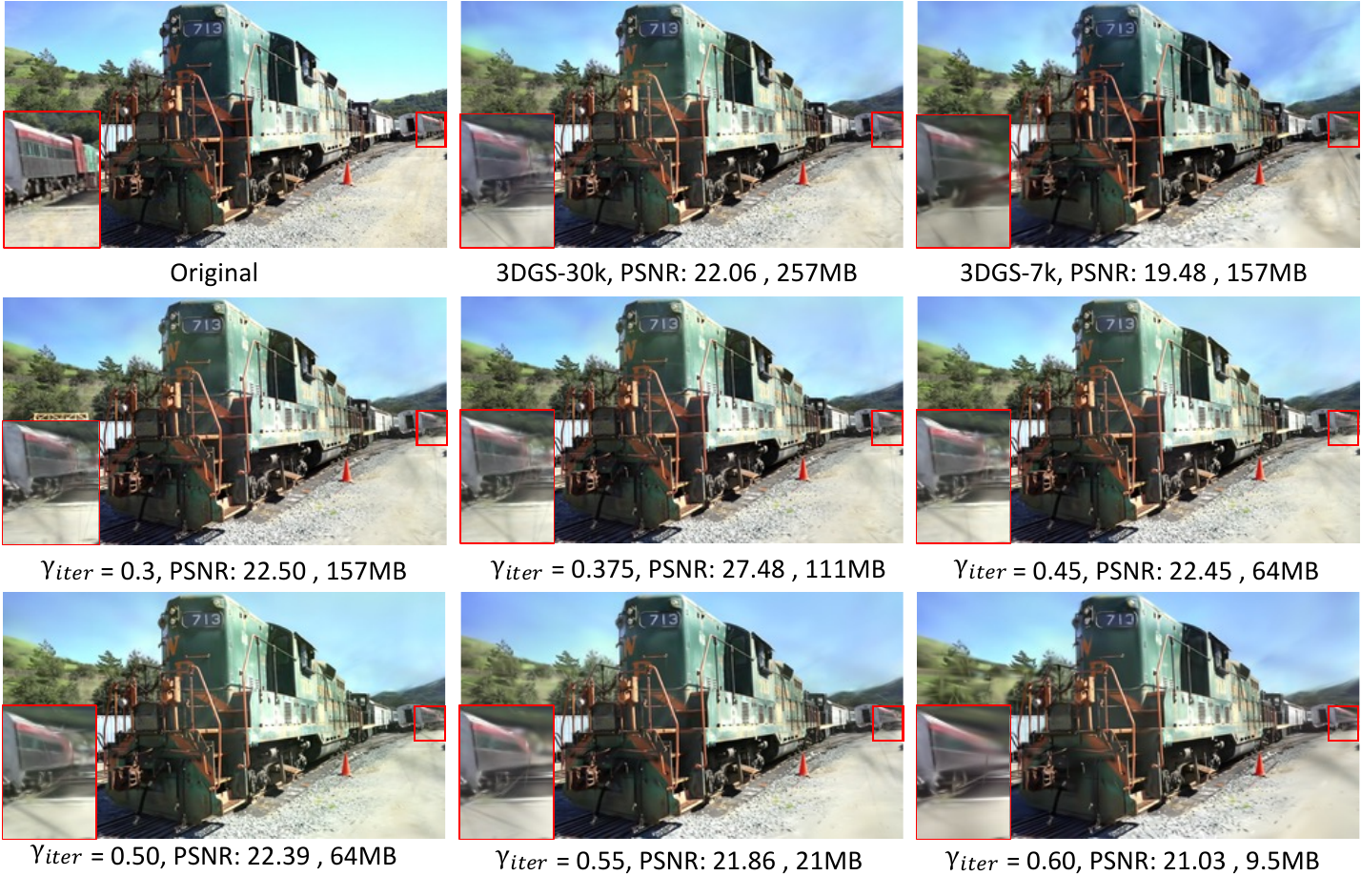}
    \centering
    \caption{Qualitative comparison of the \textit{train} scene at various pruning levels, defined by $\gamma_{\text{iter}}$ using gradient-aware iterative pruning. Our proposed method demonstrates substantially higher compression rates compared to both baselines while maintaining similar visual quality.}
    \label{fig: Pruning-Comparison-3}
\end{figure*}

We present visualizations of \textit{train} scene from the Tanks\&Temples dataset and \textit{playroom} scene from the Deep Blending dataset, all of which require substantial memory resources on average. In Fig~\ref{fig: Pruning-Comparison-2} and~\ref{fig: Pruning-Comparison-3}, we illustrate the visualizations of test set images at various pruning levels indicated by $\gamma_{\text{iter}}$. Our "trimming the fat" iterative pruning pipeline achieves noteworthy compression rates while maintaining comparable visual quality. Across all scenes depicted in the Fig.~\ref{fig: Pruning-Comparison-2} and~\ref{fig: Pruning-Comparison-3}, our method compresses the Gaussian splats by approximately $4\times$ with visual quality similar to 3DGS-30K. Furthermore, with $\gamma=0.60$, our method achieves an average compression ratio of approximately 12$\times$ while preserving visual quality comparable to 3DGS-7K.

\section{Additional Ablation Studies}

\subsection{Lottery Ticket for the Gaussian Splats?}
 We investigated the potential presence of a ``lottery ticket'' phenomenon~\cite{frankle2018lottery} for Gaussian splats. To test this hypothesis, we took an already pruned set of Gaussian splats from the Tanks\&Temples dataset and randomly reinitialized all learnable features, including spherical harmonics (SH) features, opacity, scale, and rotation. Subsequently, we attempted to train these Gaussian splats for 30,000 iterations, but they failed to converge. This experiment underscores the necessity of having a learned 3D prior to which redundant information can be pruned. It highlights the difficulty of training Gaussian splats with the minimum number of Gaussians without any prior information from the 3D scene.

\subsection{Trimming the Fat vs Compact3D~\cite{lee2023compact}.} 
\begin{figure*}[t]
\includegraphics[width=0.6\linewidth]{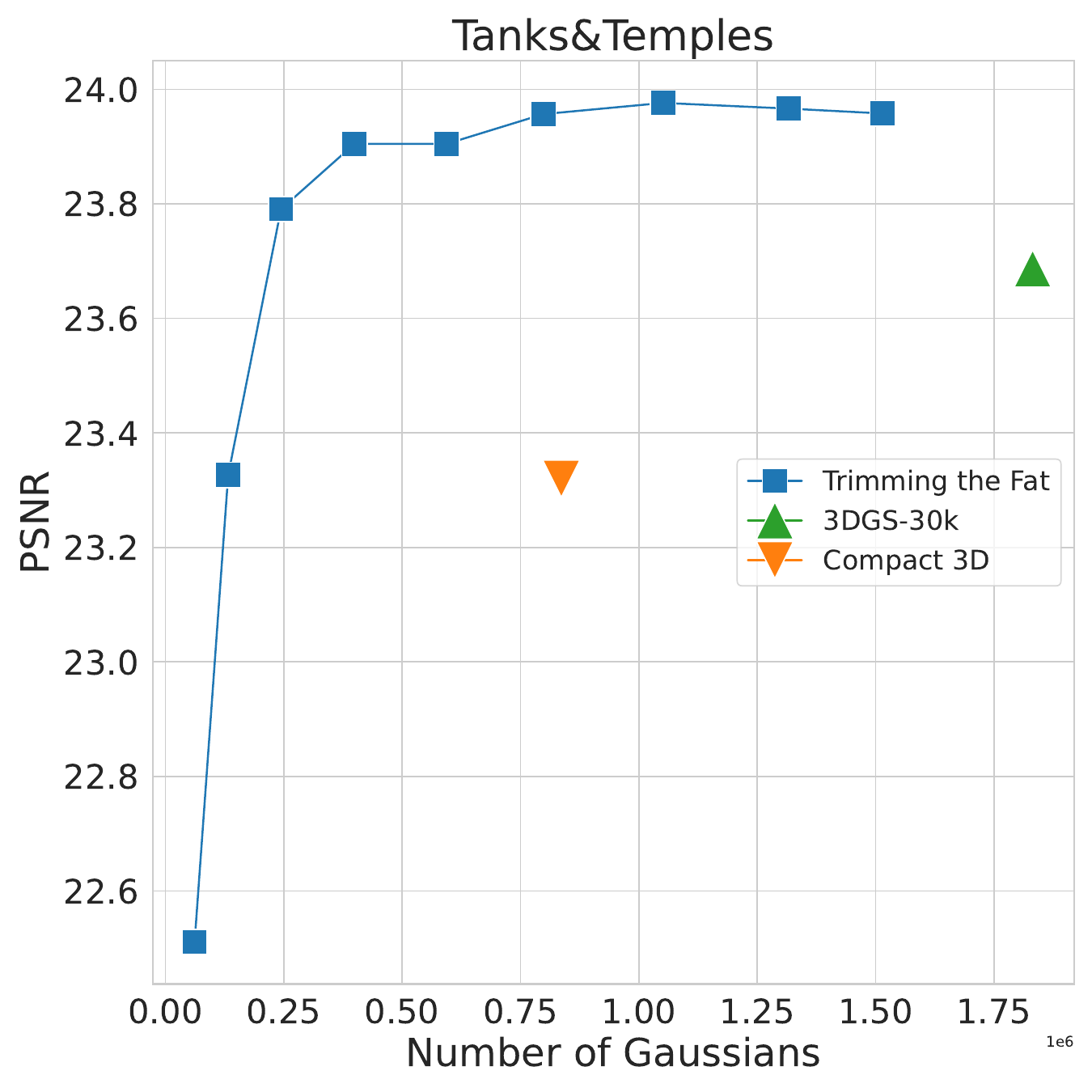}
    \centering
    \caption{The graph illustrates the performance-size trade-off achieved by our method compared to the pruning approach proposed in Compact 3D~\cite{lee2023compact} on the Tanks\&Temples dataset. }
    \label{fig: ours vs compact}
\end{figure*}

Fig.~\ref{fig: ours vs compact} depicts a comparison of PSNR and Gaussian counts between our proposed approach and Compact 3D using the Tanks\&Temples dataset. The results unequivocally highlight the superior performance of our method, demonstrating its capability to significantly reduce the number of Gaussians while maintaining baseline performance levels. These findings emphasize the effectiveness of our pruning technique and its potential to advance or replace existing compression methodologies for 3DGS.

\begin{table*}
    
    \centering
    \resizebox{\textwidth}{!}{
    \begin{tabular}{l|c|cccc|cccc|cccc}
    \toprule
      &   & \multicolumn{4}{c}{Mip-NeRF360} & \multicolumn{4}{c}{Tanks\&Temples} & \multicolumn{4}{c}{Deep Blending} \\
    \textbf{Model} & \textbf{$\gamma_{\text{iter}}$} & \textbf{SSIM$^\uparrow$} & \textbf{PSNR$^\uparrow$} & \textbf{LPIPS$^\downarrow$} & \textbf{Mem}$^\downarrow$ & \textbf{SSIM$^\uparrow$} & \textbf{PSNR$^\uparrow$} & \textbf{LPIPS$^\downarrow$} & \textbf{Mem}$^\downarrow$ & \textbf{SSIM$^\uparrow$} & \textbf{PSNR$^\uparrow$} & \textbf{LPIPS$^\downarrow$} & \textbf{Mem}$^\downarrow$ \\
    \midrule
    
   3DGS-7k$^\dag$& -   & 0.770 & 25.60 & 0.279 & 523.00 &0.767 & 21.20 & 0.280 & 270.00 & 0.875 & 27.78 & 0.317 & 386.00 \\
    3DGS-30k$^\dag$& -     & 0.815 & 27.21 & 0.214 &    734.00   & 0.841 & 23.14 & 0.183 &  411.00  & 0.903 & 29.41 & 0.243 & 676.00 \\
    3DGS-7k$^*$ & -    & 0.765 & 25.88 & 0.288 & 541.70  & 0.777 & 21.66 & 0.266 & 298.00 &  0.876 & 28.26 & 0.312 & 410.50  \\
    3DGS-30k$^*$ & -     & 0.812 & 27.46 & 0.221 & 763.40  & 0.845 & 23.69 & 0.178 & 435.50 & 0.899 & 29.46 & 0.246 & 664.50 \\
\midrule
    \multirow{8}{*}{\begin{tabular}{@{}p{2cm}@{}}3DGS-Opacity Based\\Pruning\end{tabular}} & 0.025 & 0.813 & 27.58 & 0.217 & 592.22 & 0.849 & 23.98 & 0.169 & 338.50 & 0.894 & 29.27 & 0.248 & 516.00 \\
    & 0.050 & 0.813 & 27.56 & 0.220 & 449.11 & 0.849 & 23.94 & 0.171 & 261.00 & 0.894 & 29.27 & 0.250 & 398.00 \\
    & 0.075 & 0.809 & 27.47 & 0.231 & 344.00 & 0.846 & 23.96 & 0.178 & 200.00 & 0.895 & 29.27 & 0.252 & 305.00 \\
    & 0.100 & 0.799 & 27.25 & 0.250 & 261.78 & 0.839 & 23.80 & 0.192 & 152.00 & 0.894 & 29.22 & 0.259 & 232.00 \\
    & 0.150 & 0.762 & 26.46 & 0.304 & 148.00 & 0.818 & 23.40 & 0.232 & 86.00 & 0.888 & 28.96 & 0.279 & 131.00 \\
    & 0.200 & 0.725 & 25.66 & 0.353 & 80.89 & 0.794 & 23.07 & 0.574 & 47.00 & 0.883 & 28.66 & 0.295 & 72.00 \\
    & 0.250 & 0.692 & 24.98 & 0.394 & 42.78 & 0.767 & 22.66 & 0.310 & 25.00 & 0.877 & 28.20 & 0.310 & 38.00 \\
    & 0.300 & 0.659 & 24.26 & 0.430 & 22.14 & 0.731 & 21.90 & 0.354 & 12.65 & 0.869 & 27.69 & 0.328 & 19.00 \\
\midrule
    \multirow{8}{*}{\begin{tabular}{@{}p{2cm}@{}}Trimming\\the Fat\end{tabular}} & 
    0.225 & 0.813 & 27.60 & 0.217 & 543.44 & 0.849 & 23.96 & 0.170 & 335.50 & 0.898 & 29.50 & 0.247 & 534.00 \\
& 0.275 & 0.814 & 27.60 & 0.219 & 464.33 & 0.849 & 23.97 & 0.171 & 280.75 & 0.898 & 29.48 & 0.247 & 481.50 \\
& 0.325 & 0.813 & 27.60 & 0.223 & 386.11 & 0.848 & 23.97 & 0.175 & 219.25 & 0.899 & 29.51 & 0.248 & 416.00 \\
& 0.375 & 0.810 & 27.57 & 0.231 & 311.78 & 0.844 & 23.94 & 0.187 & 153.00 & 0.899 & 29.53 & 0.250 & 339.00 \\
& 0.450 & 0.797 & 27.35 & 0.256 & 194.22 & 0.829 & 23.85 & 0.220 & 76.75 & 0.899 & 29.55 & 0.253 & 205.50 \\
& 0.500 & 0.776 & 26.93 & 0.288 & 119.44 & 0.810 & 23.56 & 0.251 & 45.25 & 0.899 & 29.52 & 0.258 & 116.00 \\
& 0.550 & 0.740 & 26.21 & 0.336 & 63.56 & 0.780 & 22.92 & 0.294 & 23.38 & 0.897 & 29.25 & 0.269 & 59.00 \\
& 0.600 & 0.690 & 25.03 & 0.394 & 29.22 & 0.761 & 22.51 & 0.319 & 14.75 & 0.887 & 28.66 & 0.293 & 25.50 \\
    \bottomrule

    \end{tabular}
    }
    \caption{Performance comparison using gradient-aware iterative pruning with different pruning levels defined by $\gamma$ against 3DGS-30k, 3DGS-7k baselines, and opacity-based iterative pruning. $^*$Reproduced using official code. $^\dag$ Reported from~\cite{kerbl20233d}. Memory size is in MBs.
    }
    \label{tab:pruning_comp}
\end{table*}

\end{document}